\documentclass{article}
\usepackage{iclr2026_conference,times}

\usepackage{hyperref}
\usepackage{url}

\PassOptionsToPackage{numbers}{natbib}

\usepackage[utf8]{inputenc} 
\usepackage[T1]{fontenc}    
\usepackage{hyperref}       
\usepackage{url}            
\usepackage{booktabs}       
\usepackage{amsfonts}       
\usepackage{nicefrac}       
\usepackage{microtype}      
\usepackage{float}
\usepackage{enumitem}

\usepackage{graphicx}
\usepackage{subcaption}
\usepackage{appendix}
\usepackage{tikz}
\usetikzlibrary{trees}
\title{Spatially-Constrained Clustering of Geospatial Features for Heat Vulnerability Assessment of Favelas in Rio de Janeiro}

\author{
  Baptiste Clemence\textsuperscript{1},
  Thomas Hallopeau\textsuperscript{1},
  Vanderlei Pascoal de Matos\textsuperscript{2},
  Laurent Demagistri\textsuperscript{1}  \&
  Joris Guerin\textsuperscript{1} \\
  \textsuperscript{1}UMR Espace-Dev, French National Research Institute for Sustainable Development (IRD), \\
  500 rue Jean-François Breton, 34393, Montpellier, France \\
  \texttt{\{baptiste.clemence,thomas.hallopeau,laurent.demagistri,joris.guerin\}@ird.fr} \\
  \textsuperscript{2}Funda\c{c}\~ao Oswaldo Cruz (Fiocruz), \\
4365 - Manguinhos, Rio de Janeiro - CEP: 21040-900, Brasil\\
  \texttt{vanderlei.pascoal@fiocruz.br}
}

\iclrfinalcopy

\begin{document}

\maketitle


\begin{abstract}
Informal settlements face disproportionate exposure to climate-related health hazards. However, existing methodologies lack systematic approaches to link diverse settlement characteristics with environmental health outcomes. We develop a data-driven framework to assess heat vulnerability in Rio de Janeiro's favelas by combining spatially-constrained clustering with land surface temperature (LST) analysis. Using remote sensing and geospatial features, we identify two distinct favela typologies: recent, well-connected settlements on flat terrain (Cluster 0) and historical, poorly-connected communities on vegetated slopes (Cluster 1). Analysis of 16 extreme heat events reveals systematic temperature differences of 2-3°C between clusters, with flat-terrain favelas experiencing significantly higher heat exposure. Our findings demonstrate that settlement morphology critically influences heat vulnerability, providing a replicable framework for targeted urban planning and public health interventions in informal settlements globally.
\end{abstract}

\vspace{-10pt}

\section{Introduction}
In 2022, the global urban population exceeded 56\% and is projected to reach 68\% by 2050~\citep{UN2024}. Informal settlements, particularly slums, house over one billion people~\citep{UNDESA2025}. They develop outside regulatory planning frameworks, often in response to rapid demographic growth, land pressure, and socio-spatial exclusion. They are characterized by high building density, precarious housing, limited access to public services, poor construction quality, increased exposure to natural hazards, and a lack of green spaces ~\citep{Kuffer2016}. For all those reasons, they present major public health challenges~\citep{Corburn2019}.

Informal settlements, called favelas in Brazil \citep{abreu1987}, exhibit significant heterogeneity in their physical characteristics, leading to different responses to health stressors such as pandemics or extreme heat events~\citep{Mendonca2017, Kaiser2025}. Understanding which settlement characteristics influence health risk exposure is essential for targeted interventions and urban planning. However, existing methodologies lack a systematic, multivariate approach to link the diverse informal settlements geospatial features with health-related environmental outcomes.

This research addresses this gap by developing a data-driven methodology to study heat vulnerability of informal settlements in Rio de Janeiro, Brazil. This is a valuable case study given the city's substantial informal settlement population and recurring public health challenges during heat waves~\citep{Peres2018}. Our approach establishes a typology of Rio's favelas through spatially-constrained clustering of remote sensing and geospatial features, including textural descriptors~\citep{demattos2021,graesser2012}, spectral indices, and OSM road network data~\citep{hallopeau2025, Owen2013}. We then analyze how the derived settlement typology relates to land surface temperature (LST) variability during extreme heat events, providing a framework for identifying heat-vulnerable favela types.

\section{Typology of Favelas}
\subsection{Data}
We use the reference data of favela boundaries from the Pereira Passos Institute (IPP, year 2022). They cover the entire metropolitan area of Rio de Janeiro.
A spatial grid with a resolution of 150 meters is generated across the entire area covered by our study, following the method proposed by \citet{hallopeau2025}, inspired by \citet{Owen2013}. The grid can be visualized in Figure~\ref{fig:carte_clusters}.

For each grid cell, we compute several descriptors from remote sensing and geospatial data, described in Table~\ref{tab:variables_favelas}. In a previous study, \citet{hallopeau2025} demonstrated the effectiveness of using these features for a \textit{favela vs. residential} cell classification task, which supports that they are relevant descriptors to establish our typology. Following \citet{hallopeau2025} methodology, we retain only grid cells with at least 70\% favela coverage to ensure the analysis focuses on areas where favelas are dominant.

As the resolution of the favela boundaries is sometimes very precise in the IPP polygons, some very close favelas are referenced as different settlements in the IPP database. We solve this issue by merging polygons less than 20 metres apart, which result in the creation of favela complexes with a single identifier.

\subsection{Method}
To establish a typology of the favelas, we aim to apply a clustering algorithm to separate the cells into unified groups of settlements presenting similar characteristics. In order to guarantee that cells belonging to the same favela are allocated to the same cluster, we must apply a constrained clustering algorithm, allowing us to incorporate must-link constraints to certain pairs of cells. To do this, we use the COP-KMeans algorithm, a variant of the classic K-Means unsupervised classification algorithm that incorporates constraints into the clustering process \citep{Wagstaff2001}. This method is particularly suitable when prior knowledge is available and the goal is to guide the automatic grouping process, while allowing part of the data to structure itself based on similarity. We use a Must-Link constraint based on the favela unique identifiers to group cells belonging to the same favela or complexes of favelas in order to try and preserve the continuity of the urban fabric.

Before training the model, NDVI, slope, and convexity are spatially aggregated to a 150-meter grid using the mean. Shannon entropy and OSM variables are computed at the same 150 m resolution. All variables are then normalized.
We selected the number of clusters (k=2) following the elbow technique and silhouette coefficient analysis~\citep{Hastie2009,Rousseeuw1987} (Figure ~\ref{fig:silhouette}).

\begin{table}[t]
\centering
\caption{Description of the variables used for the favela typology}
\label{tab:variables_favelas}
\begin{tabular}{|p{2.7cm}|p{6.5cm}|p{3.7cm}|}
\hline
\textbf{Variable} &
\textbf{Description} &
\textbf{Source} \\ \hline
NDVI &
Normalized Difference Vegetation Index. &
S2 (16/08/2022) \\ \hline
Entropy &
Measure of disorder or textural diversity &
S2 (16/08/2022) \\ \hline
Slope &
Average slope, in degrees. &
DEM GLO-30 ESA 2019 \\ \hline
Convexity &
Representing terrain
curvature &
DEM GLO-30 ESA 2019 \\ \hline
Nodes &
Number of nodes in the road network. &
OSM 2025 \\ \hline
Road length &
Total length of road segments, in meters. &
OSM 2025 \\ \hline
Mean connections &
Average connectivity of road nodes &
OSM 2025 \\ \hline
Min connections &
Min connectivity value of nodes in the cell. &
OSM 2025 \\ \hline
Max connections &
Max connectivity value of nodes in the cell. &
OSM 2025 \\ \hline
\end{tabular}
\end{table}

\subsection{Results}

\begin{figure}[t]
    \centering
    \includegraphics[width=0.9\textwidth]{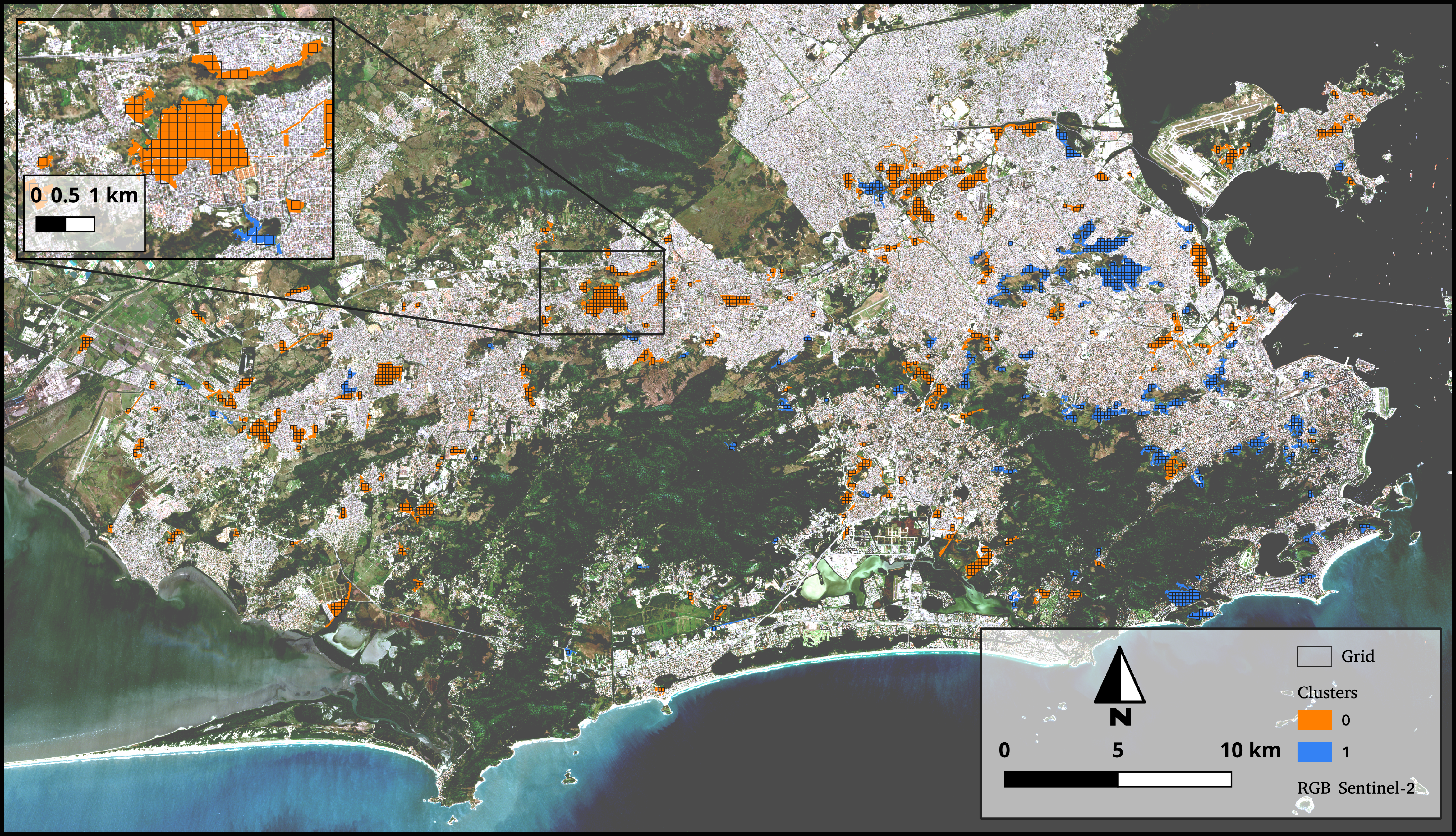}
    \caption{Spatial distribution of favela clusters identified by COP-KMeans}
    \label{fig:carte_clusters}
\end{figure}

Figure~\ref{fig:carte_clusters} shows the spatial distribution of the two identified clusters.
Cluster 1 is mainly concentrated in the city center (eastern part of the city), while Cluster 0 extends towards the west.
This spatial division reflects contrasting urban dynamics linked to the history and topography of Rio de Janeiro~\citep{abreu1987}.

\begin{figure}[t]
    \centering
    \includegraphics[width=0.9\textwidth]{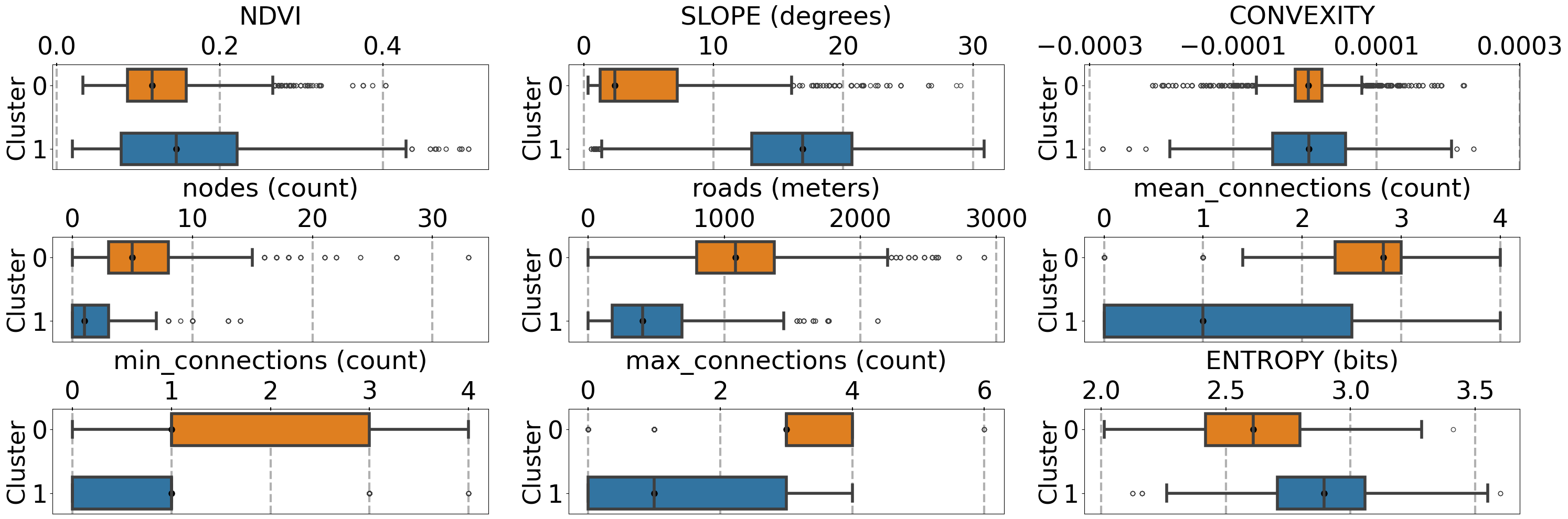}
    \caption{Distribution of explanatory variables by cluster before normalization}
    \label{fig:box_plot_variable}
\end{figure}

Figure~\ref{fig:box_plot_variable} highlights the differences in explanatory variable values between the two clusters. 
The key differentiating variables are: (1) The slope, with a median slope of 17$^\circ$ in Cluster 1 and only 2$^\circ$ in Cluster 0. (2) The road network, which is much less dense in Cluster 1, with a median road length of 400 meters per cell, compared to 1100 meters in Cluster 0. The number of road nodes is also lower in Cluster 1, with a median of 1, compared to 5 in Cluster 0, indicating a less interconnected network. Other variables present similar distributions and are not differentiating factors.

\section{Heat Vulnerability Analysis}
\subsection{Data and Methods}

We extract LST data directly from Landsat 8 and 9, Collection 2, Level-2 products, specifically using the thermal band \texttt{ST\_B10}. From an initial dataset of 40 images acquired during warm seasons between 2014 and 2025, only the 16 dates where the median LST of the entire image was $\geq 40^\circ$C under clear-sky conditions were retained. This selection criterion ensures that the analysis focuses on the hottest days, which are likely to correspond to heat-wave events. We then compare the distribution of LST between each clusters, for each of the 16 dates.

\subsection{Results}
\begin{figure}[H]
    \centering
    \includegraphics[width=0.91\textwidth]{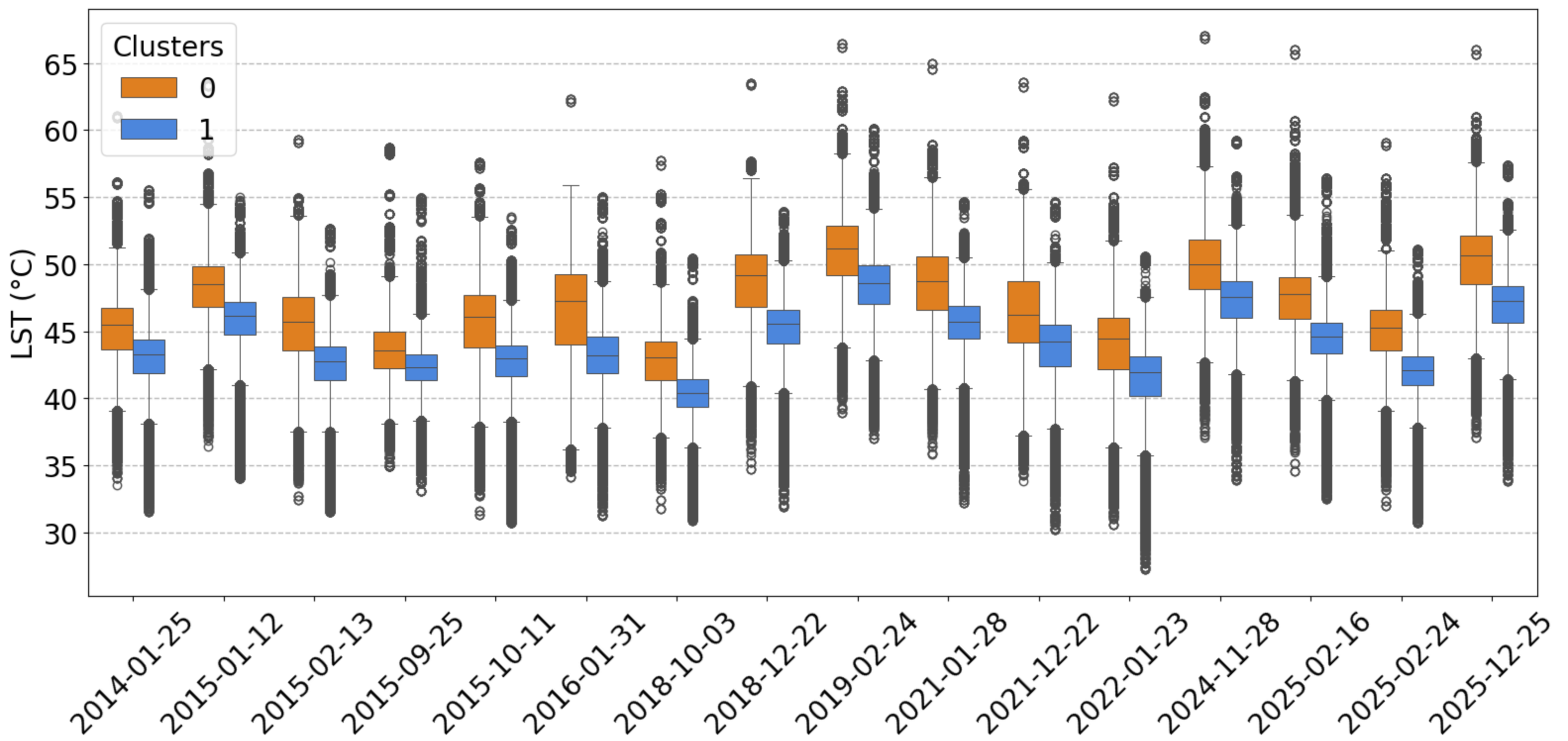}
    \caption{Comparison of LST distribution between clusters during extreme heat events.}
    \label{fig:box_plot_lst}
\end{figure}

The comparison of clusters' responses during extreme heat events is presented in Figure~\ref{fig:box_plot_lst}. The box plots show a difference in LST between clusters, regardless of the selected date. LST values of favelas are 2 to 3°C higher in Cluster 0 compared to Cluster 1. While the temperature distributions overlap, 75\% of temperature pixels in Cluster 1 remain cooler than the median of Cluster 0.

\section{Discussion}

This study reveals two distinct profiles of favelas : 
\begin{itemize}[leftmargin=15pt]
    \item Cluster 0 includes relatively recent favelas located on flat terrain. These areas are more organized, with a denser road network. Their urban structure, characterized by a high presence of artificial surfaces (buildings, roads) and bare soil, promotes increased absorption of solar radiation. This results in higher LSTs compared to the other cluster, as urban materials store and re-emit more heat than vegetated surfaces \citep{li2020}.

    \item Cluster 1 consists of the oldest favelas, historically established on vegetated slopes. These areas exhibit a slightly higher NDVI, indicating a greater presence of vegetation. LSTs are systematically cooler in this cluster, across all selected dates. This finding aligns with \citet{alves2017}, who emphasized that vegetated slopes benefit from the cooling effect of evapotranspiration and shading, thereby reducing heat absorption.
\end{itemize}

\section{Limitations and Future Work}

While our binary typology effectively captures the primary distinction between historical hillside settlements and recent flat-terrain favelas, we acknowledge several limitations that open avenues for future research.

\textbf{Typology refinement.} The choice of k=2 clusters, though supported by silhouette and elbow analyses, may oversimplify the heterogeneity within Rio's informal settlements. Future work could incorporate complementary geospatial features such as building height, building density per cell, or roof material characteristics to refine the typology.

\textbf{Spatial and temporal resolution.} The 150-meter resolution may not capture fine-scale intra-favela variations. On the other hand, our daytime LST focus might sometimes reflect material heat storage rather than true air temperature. Higher-resolution data and nighttime Landsat acquisitions would better characterize urban heat island intensity across typologies.

\textbf{Integration with health outcomes.} Linking our favela typology with epidemiological data, such as mortality during the COVID-19 pandemic ~\citep{gracie2021} or excess deaths rates during heat waves \citep{Monteiro2024}, would provide direct evidence of the health implications of different settlement types. Such integration would strengthen the policy relevance of our approach and enable targeted health protection strategies.

Despite these limitations, our methodology provides a replicable and scalable framework for linking settlement typology to urban environment health outcomes, with potential applications to other urban contexts and health-related vulnerabilities in informal settlements worldwide.

\newpage
\small
\bibliographystyle{IEEEtranN}
{\bibliography{ccai}}

\appendix 
\renewcommand{\thefigure}{A.\arabic{figure}} 
\setcounter{figure}{0} 

\section{Optimal number of clusters} \label{appendix:tree}

\begin{figure}[H]
    \centering
    \includegraphics[width=0.8\textwidth]{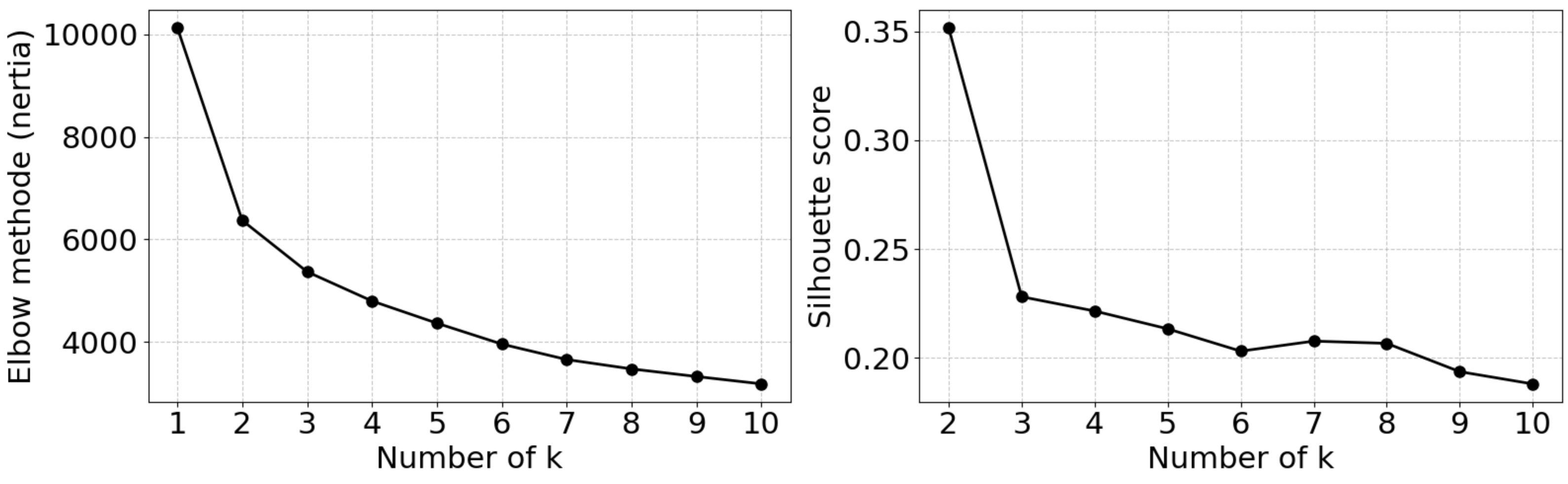}
    \caption{{Elbow method and Silhouette score}}
    \label{fig:silhouette}
\end{figure}

The selection of the number of clusters is performed with elbow method and silhouette analysis Figure~\ref{fig:silhouette}. 
The decrease in inertia in the elbow method suggests 2 to 3 clusters.
The absence of a local maximum on the silhouette curve seems to indicate uniformly distributed data, data with little variance or the need to explore other metrics. 
We choose a 2-cluster partition in this study as a first approach, which already allows us to identify a typology.

\end{document}